\documentclass{article}

\usepackage{arxiv}

\usepackage[utf8]{inputenc} 
\usepackage[T1]{fontenc}    
\usepackage{hyperref}       
\usepackage{url}            
\usepackage{booktabs}       
\usepackage{amsfonts}       
\usepackage{nicefrac}       
\usepackage{microtype}      
\usepackage{lipsum}
\usepackage{graphicx}
\usepackage{natbib}
\usepackage{times}
\usepackage{fancyvrb}
\usepackage{latexsym}
\usepackage{orcidlink} 
\usepackage{caption}
\usepackage[most]{tcolorbox}

\graphicspath{ {./images/} }

\title{Improving Estonian Text Simplification through Pretrained Language Models and Custom Datasets}

\author{
  Eduard Barbu\orcidlink{0000-0001-5414-6089} \\
  Institute of Computer Science\\
  University of Tartu\\
  Tartu, Estonia \\
  \texttt{eduard.barbu@ut.ee} \\
   \And
Meeri-Ly Muru \\
  Institute of Computer Science\\
  University of Tartu\\
  Tartu, Estonia \\
  \texttt{meerimuru@gmail.com} \\
  \And
 Sten Marcus Malva\\
  Institute of Computer Science\\
  University of Tartu\\
  Tartu, Estonia\\
  \texttt{sten.malva@gmail.com} \\
}

\begin{document}
\maketitle
\begin{abstract}
This paper presents a method for text simplification based on two neural architectures: a neural machine translation (NMT) model and a fine-tuned large language model (LLaMA). Given the scarcity of existing resources for Estonian, a new dataset was created by combining manually translated corpora with GPT-4.0-generated simplifications. OpenNMT was selected as a representative NMT-based system, while LLaMA was fine-tuned on the constructed dataset. Evaluation shows LLaMA outperforms OpenNMT in grammaticality, readability, and meaning preservation. These results underscore the effectiveness of large language models for text simplification in low-resource language settings. The complete dataset, fine-tuning scripts, and evaluation pipeline are provided in a publicly accessible supplementary package to support reproducibility and adaptation to other languages.
\end{abstract}

\keywords{text simplification \and LLM \and machine translation}

\section{Introduction}

Text simplification converts complex text into simpler forms while preserving meaning. Typical operations include sentence splitting, shortening, lexical substitution, and syntactic simplification. Simplified texts help language learners, people with cognitive impairments, and readers with low literacy, and are useful in education and public communication \cite{alva-manchego-etal-2020-data,Shardlow2014}.

Automatic text simplification (ATS) uses NLP models to perform these edits. Proprietary LLMs (e.g., GPT‑4.0, Claude) can do zero/few‑shot simplification, but often fail to reach the lowest readability levels while preserving meaning \cite{barayan2025analysing} and raise concerns about transparency and long‑term reproducibility.

Open, language‑specific systems are an alternative: they allow control over training data and domain goals, and avoid dependence on commercial APIs—especially important for low‑resource languages like Estonian.

We study two architectures for Estonian ATS: OpenNMT (simplification as monolingual MT) and a fine-tuned LLaMA model trained on a newly built Estonian Simplification Dataset. We complement automatic evaluation using BLEU and SARI scores with human evaluation by native speakers.
\textbf{Contributions.}
\begin{itemize}
\setlength{\itemsep}{2pt}\setlength{\parskip}{0pt}\setlength{\parsep}{0pt}
\item A new Estonian Simplification Dataset combining translated, GPT‑4.0–generated, and manually validated pairs.
\item A head‑to‑head comparison of OpenNMT and fine‑tuned LLaMA for Estonian.
\item Evidence from human ratings that LLaMA trained on Estonian data outperforms OpenNMT.
\item Released scripts/configs for easy adaptation to other low‑resource languages.
\end{itemize}

The remainder of the paper is organized as follows. Section~\ref{ESD} describes the Estonian simplification dataset, Section~\ref{benchmarking} benchmarks English models and motivates OpenNMT as a baseline, Section~\ref{LLaMA} details LLaMA fine-tuning, and the final sections present evaluation and conclusions.

\section{Related Work}

Research on ATS has been dominated by English, with surveys charting the progression from handcrafted rules to neural models and large-scale pretraining \cite{saggion2017automatic, espinosa-zaragoza-etal-2023-review, Shardlow2014ASO, paetzold2016survey}. 

\textbf{Rule-based approaches.} Early work relied on handcrafted rules for lexical substitution and grammatical restructuring. Foundational systems by \citet{chandrasekar-etal-1996-motivations} and \citet{10.3115/1118984.1118986} introduced syntactic transformations to aid reading comprehension. These methods were interpretable but brittle and difficult to scale across domains.

\textbf{User-group oriented projects.} Several initiatives highlighted the value of tailoring simplification to specific audiences. The HAPPI project simplified dialogue systems for people with aphasia \cite{10.1145/1168987.1169027}. PorSimples targeted Brazilian Portuguese for readers with low literacy \cite{aluisio-etal-2010-readability}. The FIRST project adapted texts for individuals with autism spectrum disorder and cognitive disabilities \cite{BARBU20155076}. These projects emphasized that simplification is not one-size-fits-all, but must reflect user needs and linguistic context.

\textbf{Neural models.} The rise of deep learning shifted ATS toward Seq2Seq architectures with attention. The DRESS model \cite{zhang-lapata-2017-sentence} pioneered reinforcement learning for simplification, optimizing grammaticality, meaning preservation, and simplicity simultaneously. Later work integrated semantic parsing \cite{zhao-etal-2018-integrating}, while controllable models introduced explicit knobs for simplification strength \cite{sheang2021controllable,martin2020controllable}. These methods improved fluency and structural edits compared to rule-based systems.

\textbf{Pretrained language models.} Transformer-based pretrained models pushed performance further. T5 \cite{10.5555/3455716.3455856} framed simplification as a text-to-text problem, enabling broad transfer. Proprietary models like GPT-4 \cite{openai2023gpt4} achieve strong zero/few-shot results. However, they remain closed-source, expensive, and less transparent. Open-access LLaMA models \cite{touvron2023llamaopenefficientfoundation} offer a promising alternative: large multilingual pretraining combined with accessibility for fine-tuning in low-resource languages.

\textbf{Datasets.} English ATS benefited from multiple corpora. For example, WikiSmall \cite{zhu-etal-2010-monolingual} aligned Wikipedia with Simple English Wikipedia. TurkCorpus \cite{xu-etal-2016-optimizing} provided multiple human simplifications per sentence, improving evaluation robustness. Newsela \cite{xu-etal-2015-problems} offered professional multi-level rewrites, though restricted to licensed access. Unsupervised methods \cite{alva2021unsupervised, jiang2020neural} explored automatic sentence alignment and pseudo-parallel generation to overcome data scarcity.

\textbf{Low-resource languages.} Outside English, ATS research is limited. . Estonian ATS has so far been confined to two undergraduate theses: one on WordNet-based lexical substitution and another on template-driven syntactic rules. No large-scale, publicly available dataset previously existed.

While ATS has matured for English, low-resource languages remain underserved. We address this gap by releasing the first large Estonian simplification dataset and benchmarking OpenNMT against fine-tuned LLaMA, showing that reproducible ATS is feasible for underrepresented languages.

\section{Building the Estonian Simplification Dataset} \label{ESD}

In this section, we describe the construction of the Estonian Simplification Dataset. The goal was to create a resource large and diverse enough to train deep neural models. Building such a dataset from scratch would have been prohibitively time-consuming, given the absence of prior Estonian simplification data.

We aligned the dataset with English corpora and established clear guidelines for Estonian-specific phenomena. This dual alignment with international simplification efforts and local linguistic norms ensures both relevance and linguistic grounding.

The first source for our dataset is the Turk corpus \cite{xu-etal-2016-optimizing}, which contains 2,359 original Wikipedia sentences, each accompanied by eight simplified versions crowdsourced via Amazon Mechanical Turk. This corpus captures a variety of simplification strategies, including lexical substitution, paraphrasing, and sentence restructuring. We manually translated relevant portions of this corpus into Estonian, but only retained sentence pairs where a clear simplification relationship was preserved after translation. In some cases, simplification evident in the English original did not carry over naturally into Estonian due to syntactic or lexical differences, and such pairs were excluded.

The second source is the Wikipedia Data Set 2.0 \cite{kauchak-2013-improving}, which includes 167,689 aligned sentence pairs between standard and Simple English Wikipedia. Small subsets of this dataset were machine-translated into Estonian and subsequently corrected by native annotators. However, despite this effort, the resulting translations were often semantically inconsistent or lacked coherence, rendering the dataset less useful for training.

To address these challenges and ensure linguistic consistency, we developed the Estonian Simplification Guidelines, partly inspired by the methodology and goals of the Newsela Corpus \cite{xu-etal-2015-problems}. Newsela includes professionally edited news articles rewritten at multiple reading levels. Although this corpus is not openly available, it provides a valuable reference for developing high-quality simplification guidelines.

The simplification guidelines consist of a detailed 10-page document that outlines specific strategies adapted to the morphosyntactic properties of Estonian. The guidelines concentrate on three main areas:
\begin{itemize}
\item \textbf{Grammatical complexity:} Simplifying inflectional morphology, verb tense usage, and mood markers.
\item \textbf{Syntactic complexity:} Splitting long or embedded clauses, simplifying coordination, and reordering sentence constituents for clarity.
\item \textbf{Lexical simplification:} Replacing low-frequency words with more common synonyms, using hypernyms, and avoiding domain-specific jargon.
\end{itemize}

Following best practices in prior English-language simplification work, the primary corpus source for large-scale simplification is Estonian Wikipedia. Articles were extracted from the latest Wikipedia dumps, segmented into sentences, and filtered to include only those longer than 15 words.

To generate a large volume of simplification examples efficiently, we employed GPT-4.0 \cite{openai2023gpt4}, a state-of-the-art proprietary language model that ranks highly on the Chatbot Arena LLM Leaderboard. Although relying on GPT-4.0 incurs some costs, these are significantly lower than the expense of hiring human annotators for full-scale manual simplification.

GPT-4.0's flexibility allows it to assume tailored roles and styles through prompting. We leveraged this feature by designing templates that embedded persona-driven behavior to guide simplification style and focus. Prior research has shown that persona-based prompting outperforms general prompting in specific tasks \cite{pataranutaporn2021ai, wang2024rolellmbenchmarkingelicitingenhancing}.

Throughout development, we tested multiple prompting strategies across GPT-4.0 instances. Initial templates used no persona framing, while later iterations embedded role-based behaviors targeting syntactic or lexical operations. After extensive experimentation, the most effective results came from sequential prompting: first performing lexical simplification, followed by syntactic restructuring. Although recent studies have questioned the general effectiveness of persona-based prompting \cite{zheng2024ahelpfulassistantreally}, in our task setting, the agent-style prompts proved superior. By agent-style prompts, we refer to role-based instructions instantiating a simplification assistant, such as: “You are a lexical simplification assistant tasked with simplifying Estonian text.”

The breakdown of simplified sentences generated using GPT-4.0 is shown in Figure~\ref{fig:simplification_breakdown}. Out of 47,112 simplified sentence pairs, 28,479 were produced using our initial generic prompt, and 18,633 were generated using agent-style prompts for lexical and syntactic tasks.

\begin{center}
\begin{minipage}{0.9\linewidth}
\centering
\vbox{
    \textbf{3,992 Articles Simplified} \\
    $\downarrow$ \\
    \textbf{47,112 Simplified Sentence Pairs} \\
    $\swarrow$ \hspace{1cm} $\searrow$ \\
    \textbf{28,479} \hspace{1cm} \textbf{18,633} \\
    $\downarrow$ \\
    \textbf{9,913 Sentence Pairs Corrected by Annotators}
}
\captionof{figure}{Breakdown of Estonian Dataset Pairs}
\label{fig:simplification_breakdown}
\end{minipage}
\end{center}

Below is an excerpt from the lexical simplification agent used in our GPT-4.0 prompts (see \ref{prompt:lexical_agent}). The agent uses a few-shot learning strategy with example-based guidance to consistently apply Estonian lexical simplification techniques.

\begin{tcolorbox}[breakable,
  title={Prompt Example: Lexical Simplification Agent},
  label=prompt:lexical_agent,
  colback=gray!5, colframe=gray!40!black, fonttitle=\bfseries,
  left=6pt,right=6pt,top=6pt,bottom=6pt]
\textbf{Instruction:} \\
You are a lexical simplification assistant tasked with simplifying Estonian text.
Your role is to receive Estonian sentences and transform them by simplifying complex words and phrases while maintaining the original meaning.

\medskip
\textbf{Lexical Simplification Guidelines:}
\begin{itemize}
\setlength{\itemsep}{2pt}\setlength{\parskip}{0pt}\setlength{\parsep}{0pt}
\item Replace difficult words with simpler, more common alternatives.
\item Carefully consider context to preserve the original meaning.
\item Avoid simplifying proper nouns, technical terms without simpler equivalents, and words essential to the sentence's meaning.
\item Remove adjectives that do not add critical meaning.
\item Use simple, common verbs.
\item Prefer simpler tenses to enhance clarity.
\end{itemize}

\medskip
\textbf{Persona:} \\
You are an expert in lexical simplification, focused on making Estonian text more accessible.
Your approach balances reducing complexity with maintaining meaning, ensuring the text remains clear and true to the original.

\medskip
\textbf{Example:} \\
Original: Epidemioloogia uurib nakkushaigusi ja nende tõkestamist. \\
Simplified: Epidemioloogia uurib nakkushaigusi ja nende peatamist.
\end{tcolorbox}

The final dataset statistics are presented in Table~\ref{tab:dataset_comp}. In total, we compiled 50,416 sentence pairs across three primary sources. This corpus forms the backbone of our experiments and is used to train the models presented in the following sections.

\begin{table}[h!]
\centering
\begin{tabular}{|l|c|}
\hline
\textbf{Source}            & \textbf{Sentence Pairs} \\ \hline
Turk                       & 1,896                    \\ \hline
Wikipedia Data Set 2.0     & 1,408                    \\ \hline
GPT-4.0                    & 47,112                   \\ \hline
\textbf{Total Pairs}       & \textbf{50,416}          \\ \hline
\end{tabular}
\caption{Composition of the Estonian Simplification Dataset}
\label{tab:dataset_comp}
\end{table}

This dataset enables the fine-tuning of deep neural network architectures tailored for Estonian text simplification, as described in the following sections.

\section{Benchmarking Simplification Models for English Language} \label{benchmarking}

To select an appropriate baseline model for Estonian text simplification, several established English simplification systems with publicly available code were benchmarked. The evaluation protocol followed the methodology outlined by \citet{zhang-lapata-2017-sentence}.

The first system evaluated applies machine translation (MT) techniques to the simplification task. While MT systems are traditionally designed to translate across languages, their sequence-to-sequence (Seq2Seq) architectures, including both recurrent and transformer-based models, can be repurposed to translate complex sentences into simpler ones in the same language.

One foundational model is built using OpenNMT \cite{klein2018opennmt}, a widely used neural MT framework. The text simplification system described in \citet{neural-text-simplification} is implemented with an encoder-decoder architecture, typically using Recurrent Neural Networks (RNNs) such as LSTMs or GRUs. An attention mechanism is integrated to help the decoder focus on the most relevant parts of the input sentence.

OpenNMT-py, the PyTorch-based implementation, was selected for its flexibility and ease of integration. The configuration used in the experiments included:
\begin{itemize}
    \item Two encoder-decoder layers with 500 hidden units per layer,
    \item A dropout rate of 0.3 to reduce overfitting,
    \item Fifteen training epochs with Stochastic Gradient Descent (SGD) and a learning rate of 0.1.
\end{itemize}

The second model evaluated is DRESS \cite{zhang-lapata-2017-sentence}, a simplification system trained using reinforcement learning. Unlike standard supervised learning approaches, DRESS optimizes simplification quality through a reward-based signal, encouraging a balance between grammaticality, meaning preservation, and simplicity. A pretrained encoder-decoder model is used, and reinforcement learning is applied to iteratively improve output quality based on the designed reward function.

The third model is T5 (Text-To-Text Transfer Transformer) \cite{raffel2020exploring}, a unified framework in which all NLP tasks are cast as text generation problems. T5 was pretrained on the C4 dataset and has demonstrated strong generalization capabilities. It uses a transformer-based encoder-decoder architecture and supports task-specific prompting.

For the simplification task, T5 was fine-tuned over five epochs using the Adam optimizer. Input sentences were prefixed with the instruction “simplify.” Control tokens encoding structural properties (e.g., character length ratio, Levenshtein distance) were incorporated to enhance simplification quality.

All models were trained and evaluated on the WikiSmall corpus \cite{zhu-etal-2010-monolingual}, which consists of aligned sentence pairs from English Wikipedia and Simple English Wikipedia. The training set includes 89,042 sentence pairs, while the test set comprises 100 pairs.

Three standard metrics were used for evaluation:

\begin{itemize}
    \item \textbf{BLEU} \cite{papineni2002bleu}, which measures n-gram overlap between system output and reference simplifications. While informative, BLEU does not reward deletion or simplification explicitly.
    \item \textbf{SARI} \cite{xu-etal-2016-optimizing}, which evaluates the system output against both input and reference sentences, rewarding appropriate additions, deletions, and retention.
    \item \textbf{FKGL} (Flesch-Kincaid Grade Level) \cite{flesch1948flesch}, which estimates readability by calculating sentence length and syllable count. Lower scores correspond to simpler and more accessible text.
\end{itemize}

Evaluation results are shown in Table \ref{tab:eng_eval}.

\begin{table}[h!]
\centering
\begin{tabular}{|l|c|c|c|}
\hline
\textbf{Model} & \textbf{BLEU} & \textbf{SARI} & \textbf{FKGL} \\ \hline
\textbf{DRESS}    & \textbf{47.93} & 13.61  & 11.35 \\ \hline
\textbf{OpenNMT}  & 44.61 & 35.82  & 9.97  \\ \hline
\textbf{T5}       & 30.88 & \textbf{41.21}  & \textbf{7.23}  \\ \hline
\end{tabular}
\caption{Evaluation of the simplification models}
\label{tab:eng_eval}
\end{table}

The results indicate that T5 achieved the highest SARI and lowest FKGL scores, suggesting superior simplification quality and readability. However, its BLEU score was lower, indicating less lexical overlap with the reference. DRESS attained the highest BLEU score but scored poorly on SARI and FKGL, reflecting a tendency to copy input sentences with minimal simplification. OpenNMT demonstrated balanced performance across all metrics, excelling particularly in structure simplification and lexical reduction.

Given its robustness and ease of deployment, OpenNMT was selected as the baseline model for Estonian text simplification, especially considering its suitability for low-resource scenarios and independence from large-scale pretraining.

\section{Simplification with Fine-Tuned LLaMA} \label{LLaMA}

LLaMA \cite{touvron2023llamaopenefficientfoundation} is a family of open-access large language models developed by Meta, designed to offer high performance while maintaining computational efficiency. Unlike some of the larger proprietary models, LLaMA models are optimized for accessibility and can be fine-tuned and deployed in resource-constrained environments. These models are trained on diverse, publicly available multilingual corpora and support a broad range of NLP tasks, including generation, summarization, translation, and simplification.

LLaMA 3.1, introduces enhanced efficiency and accuracy compared to previous versions. It is available in various parameter configurations, with sizes up to 65 billion. The training data includes multiple languages, and while the exact sources for Estonian remain undisclosed, it is known that Estonian was included in the pretraining corpus. This multilingual pretraining provides a suitable foundation for adapting the model to Estonian-specific tasks.

To enable sentence-level simplification in Estonian, both LLaMA 3.0 and LLaMA 3.1 with 8 billion parameters were fine-tuned using the Estonian Simplification Dataset (Section~\ref{ESD}). The fine-tuning process was organized into the following steps:

\begin{enumerate}
    \item \textbf{Pretrained Model Initialization}: The LLaMA models and tokenizer were loaded from their base checkpoints. Due to their prior exposure to multilingual content, these models serve as strong initializations for low-resource language tasks such as Estonian simplification.
    
    \item \textbf{Integration of the Unsloth Library}: The Unsloth library\footnote{\url{https://github.com/unslothai/unsloth}} was used to optimize fine-tuning. This framework leverages Low-Rank Adaptation (LoRA) and QLoRA to reduce memory consumption and computational cost. It also supports mixed-precision training and efficient memory allocation, making it well suited for fine-tuning LLaMA models on modern GPUs.
    
    \item \textbf{Training Configuration}: Hyperparameters were adjusted to balance efficiency and convergence. Key parameters included learning rate, batch size, and total number of steps. Mixed precision (FP16) and gradient accumulation were employed to stabilize training. The resulting model learned to generate simplified Estonian sentences while preserving meaning and grammaticality.
\end{enumerate}

After training, both the fine-tuned model and tokenizer were serialized for downstream use in inference pipelines. The resulting model serves as an Estonian-specific simplification engine that is fully self-hosted and independent of external proprietary APIs.

\section{Evaluation} \label{evaluation}

The evaluation of the Estonian simplification models was conducted using a combination of automatic metrics and human judgments. This dual approach ensures quantitative and qualitative assessment, since automatic metrics alone often miss fluency, meaning, and simplification quality.

A subset of 100 complex sentences was randomly sampled from the Estonian Simplification Dataset. These sentences were manually simplified by two native Estonian linguists to form a high-quality gold standard. The resulting parallel dataset was then used to evaluate the performance of both the OpenNMT and LLaMA-based models.

\subsection{Automatic Evaluation} \label{automatic}

To address the scarcity of Estonian simplification data, the English WikiSmall corpus was translated into Estonian using the Tartu NLP machine translation service \cite{korotkova-fishel-2024-estonian}. This enabled training the OpenNMT model entirely on Estonian text, albeit derived from English simplification pairs, resulting in a larger and more diverse dataset than the Estonian Simplification Dataset.

For LLaMA, both the 3.0 and 3.1 versions were fine-tuned on the Estonian Simplification Dataset, excluding the 100-sentence test set. Qualitative analysis showed that LLaMA 3.1 consistently outperformed its predecessor, and it was therefore selected for final evaluation.

Training for OpenNMT was performed using the OpenNMT-py library with a shared vocabulary of 60,000 tokens. Vocabulary sharing was enabled to improve token alignment between source and target. The training process saved 12 checkpoints, and the best-performing checkpoint (based on dev loss) was selected to avoid overfitting.

For LLaMA 3.1, fine-tuning was conducted over 500 training steps, with a warmup of 100 steps. A batch size of 8 with gradient accumulation over 2 steps was used. The AdamW optimizer was configured with a learning rate of 0.00005, weight decay of 0.01, and cosine learning rate scheduling. Mixed precision (FP16) training was used to accelerate computation and reduce memory usage. A fixed seed (42) was applied for reproducibility. The resulting model was serialized and deployed for testing.

\begin{figure}[h!]
    \centering
    \includegraphics[width=0.45\textwidth]{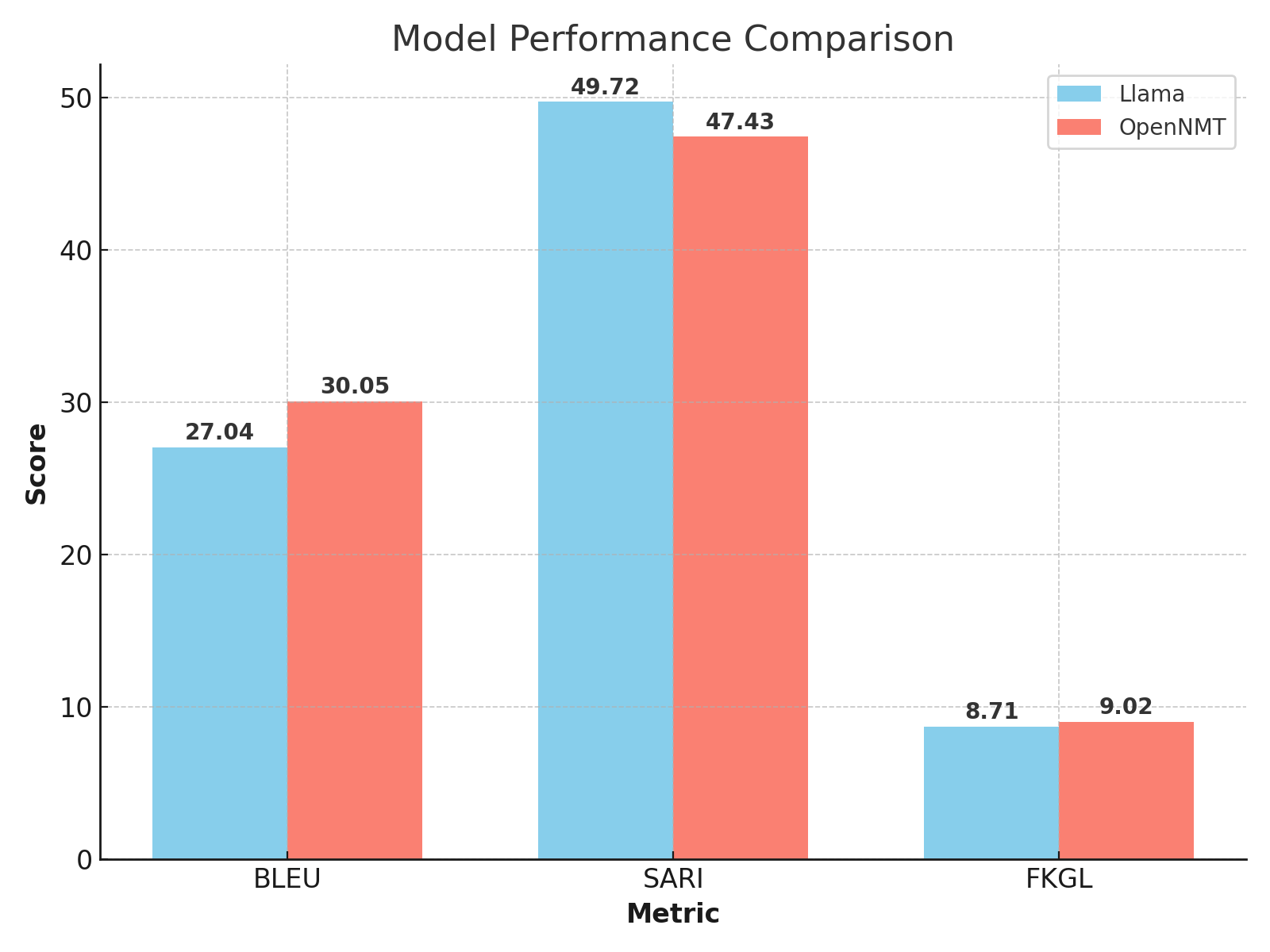}
    \caption{Comparison of Model Performance on BLEU, SARI, and FKGL metrics for LLaMA 3.1 and OpenNMT}
    \label{fig:model_performance}
\end{figure}

Evaluation was carried out using three standard metrics:

\begin{itemize}
    \item \textbf{BLEU} \cite{papineni2002bleu}: Measures n-gram overlap between model output and reference simplifications. It rewards fluency but penalizes deletion.
    \item \textbf{SARI} \cite{xu-etal-2016-optimizing}: Designed specifically for simplification, this metric compares system output to both the input and reference, rewarding appropriate edits.
    \item \textbf{FKGL} \cite{flesch1948flesch}: The Flesch-Kincaid Grade Level estimates the readability of the generated output.
\end{itemize}

As shown in Figure~\ref{fig:model_performance}, OpenNMT achieved a higher BLEU score (30.05 vs. 27.04), indicating stronger n-gram similarity with the reference texts. However, LLaMA 3.1 outperformed OpenNMT on SARI (49.72 vs. 47.43), suggesting it made more effective simplification edits. The FKGL scores (8.71 for LLaMA 3.1 vs. 9.02 for OpenNMT) show that both systems improved readability, with a slight advantage for LLaMA. These mixed results highlight the limitations of relying solely on automated metrics.

\subsection{Manual Evaluation} \label{manual}

To complement the automatic evaluation, a human assessment was conducted by two independent native Estonian linguists. Fifty randomly selected sentence pairs, simplified by each model, were rated across four distinct dimensions: grammaticality (G), readability (R), preservation of meaning (M), and reduction effort (R\textsubscript{eff}). Each criterion was scored on a 0–4 scale, with 4 denoting excellent performance and 0 indicating a complete failure. The rating guidelines were collaboratively developed and piloted to ensure reliability. Annotator disagreements were resolved through discussion to produce consensus scores.

\vspace{1em}
\noindent\textbf{Grammaticality (G)} assesses the syntactic correctness of the output. A score of 4 corresponds to a grammatically flawless sentence with native-level fluency. Lower scores indicate increasingly frequent or severe errors, including agreement mismatches, fragmentary structures, or incoherent phrasing.

\vspace{0.5em}
\noindent\textbf{Readability (R)} measures how easy the simplified sentence is to comprehend. This includes factors such as sentence length, lexical familiarity, and word order clarity. A top score of 4 reflects a well-structured, fluent sentence that is easy to read without effort.

\vspace{0.5em}
\noindent\textbf{Preservation of Meaning (M)} captures the extent to which the simplified sentence retains the key information and intent of the original. A score of 4 indicates full semantic preservation, while lower scores reflect omission, distortion, or misinterpretation.

\vspace{0.5em}
\noindent\textbf{Reduction Effort (R\textsubscript{eff})} refers specifically to the degree of surface-level simplification applied—such as shortening, word substitution, and structural flattening. It does not measure quality or appropriateness in isolation but indicates whether any simplification effort was actually made. This criterion is separated from fluency and faithfulness to avoid conflating simplification activity with overall simplification quality. It is important to note that sentence simplification as a task ideally balances all four criteria rather than optimizing for reduction alone.

\begin{table}[h!]
\centering
\begin{tabular}{|l|c|c|c|c|c|}
\hline
\textbf{Model} & \textbf{G} & \textbf{R} & \textbf{M} & \textbf{R\textsubscript{eff}} & \textbf{Overall} \\ \hline
\textbf{LLaMA 3.1} & \textbf{3.46} & \textbf{3.26} & \textbf{3.24} & \textbf{2.16} & \textbf{3.03} \\ \hline
\textbf{OpenNMT}   & 2.26 & 2.04 & 1.76 & 0.94 & 1.60 \\ \hline
\end{tabular}
\caption{Mean human ratings: Grammaticality (G), Readability (R), Meaning (M), Reduction Effort (R\textsubscript{eff}), and Overall Average}
\label{tab:manual_eval}
\end{table}

\vspace{1em}
The results in Table~\ref{tab:manual_eval} indicate that LLaMA 3.1 consistently outperformed OpenNMT across all dimensions. The largest gap appeared in preservation of meaning (3.24 vs. 1.76), confirming that LLaMA maintained semantic integrity more reliably. It also exhibited significantly higher grammatical fluency and readability. Although both models attempted simplification, LLaMA applied more effective and deliberate reduction strategies (2.16 vs. 0.94), including sentence segmentation, lexical replacement, and omission of unnecessary modifiers.

These findings reinforce the importance of human evaluation in ATS research. While automatic metrics provide useful benchmarks, they often fail to capture deeper qualitative aspects—such as nuance, clarity, or subtle meaning loss—that are crucial for practical deployment in real-world, user-sensitive applications.

\section{Conclusions}

This study explored two approaches to Estonian text simplification: a neural machine translation model using OpenNMT and a fine-tuned large language model, LLaMA. Given the limited resources for Estonian ATS, we created the Estonian Simplification Dataset by combining translated data and GPT-4.0-generated simplifications. The experimental results show that the LLaMA model, fine-tuned on this dataset, consistently outperforms OpenNMT across key criteria, including readability, grammaticality, and meaning preservation.

The evaluation showed that standard metrics, such as BLEU and SARI, were insufficient to determine a clear winner between OpenNMT and LLaMA. While BLEU scores marginally favored OpenNMT, reflecting closer alignment with reference texts, SARI scores suggested that LLaMA might better capture the simplification process by adding, deleting, or altering content for readability. However, neither metric alone fully encapsulated critical aspects of simplification quality, such as meaning preservation and readability. These findings underscore the limitations of automated metrics and the necessity of manual evaluation. Despite these limitations, BLEU and SARI are widely adopted in ATS research, enabling comparison with prior work.

The manual evaluations highlighted LLaMA's superior performance, particularly in maintaining the original meaning and applying effective simplification techniques, such as sentence splitting and lexical substitution. These findings underscore the potential of LLMs for handling low-resource languages, with fine-tuning proving effective in adapting pre-trained models to the specific linguistic and structural features of Estonian.

Importantly, the fine-tuning methodology presented in this work is not limited to Estonian. The approach, based on openly available LLaMA models and lightweight tuning via the Unsloth framework, can be readily applied to other low-resource languages with comparable syntactic or morphological complexity. By adapting the data collection, annotation guidelines, and persona-based prompting strategies introduced here, researchers can build domain- and language-specific simplification systems tailored to their needs. To support this, we share scripts, configuration templates, and model inference examples in the supplementary section to facilitate replication and reuse in future work.

This research contributes to the underexplored area of Estonian ATS, demonstrating that LLMs, supported by targeted persona prompting and data resources, can achieve meaningful simplifications in a low-resource language. 
It also supports practical applications, such as creating educational tools and improving accessibility for Estonian speakers with cognitive disabilities, aligning with user groups like those in the HAPPI, FIRST, and CLEAR projects. Future work will focus on expanding the dataset and incorporating additional human corrections of the simplified sentences. Furthermore, this research lays the groundwork for exploring document-level simplification for Estonian.

\section{Supplementary Material}

We release all datasets, models, and tools developed in this study to support reproducibility:

\begin{itemize}
    \item \textbf{Dataset:} Estonian Text Simplification Dataset (50,416 sentence pairs), combining GPT-4.0 generations, manual corrections, Wikipedia data, and machine-translated English corpora. \\
    \url{https://huggingface.co/datasets/vulturuldemare/Estonian-Text-Simplification}

    \item \textbf{Models:} 
    \begin{itemize}
        \item Fine-tuned LLaMA 3.1: \url{https://huggingface.co/datasets/vulturuldemare/Estonian-Text-Simplification/resolve/main/llama31-model.zip}
        \item OpenNMT model: \url{https://huggingface.co/datasets/vulturuldemare/Estonian-Text-Simplification/resolve/main/openNMT-SimplificationModel.pt}
    \end{itemize}

    \item \textbf{Applications:} 
    \begin{itemize}
        \item LLaMA-based web app: \url{https://github.com/SoimulPatriei/webapp-llama}
        \item OpenNMT-based web app: \url{https://github.com/SoimulPatriei/webapp-opennmt}
    \end{itemize}
\end{itemize}

All resources are openly available under permissive licenses.

\section*{Acknowledgments}

This work was supported by the EKTB55 project \emph{Text Simplification for Estonian} and by PRG2006 \emph{Language Technology for Low-Resource Finno-Ugric Languages and Dialects}. We thank the annotators who contributed to the Estonian Simplification Dataset and the three anonymous reviewers for their valuable feedback.

\bibliographystyle{plainnat} 
\bibliography{estonian-simplification} 

\end{document}